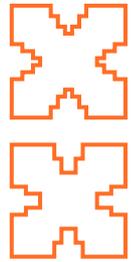

# Amplifying The Uncanny

**Terence Broad**
t.broad@gold.ac.uk
Goldsmiths, University of London,
United Kingdom

**Frederic Fol Leymarie**
ffl@gold.ac.uk
Goldsmiths, University of London,
United Kingdom

**Mick Grierson**
m.grierson@arts.ac.uk
Creative Computing Institute, University of
the Arts London, United Kingdom

**Deep neural networks have become remarkably good at producing realistic deepfakes, images of people that (to the untrained eye) are indistinguishable from real images. Deepfakes are produced by algorithms that learn to distinguish between real and fake images and are optimised to generate samples that the system deems realistic. This paper, and the resulting series of artworks *Being Foiled* explore the aesthetic outcome of inverting this process, instead optimising the system to generate images that it predicts as being fake. This maximises the unlikelihood of the data and in turn, amplifies the uncanny nature of these machine hallucinations.**

## 1. Introduction

In recent years, machine learning systems have become remarkably good at producing images, most notably of human faces, that can trick the human eye into believing they are real. Well documented controversy surrounds the potential consequences of deepfakes. Although, usingdeepfakes to spread disinformation in the political sphere has not yet occured in the way many predicted—and may instead be a convenient excuse for political actors to discredit the veracity of inconvenient footage (Breland 2019). However, deepfakes are disturbingly prevalent in pornographic websites where the image of a person (almost exclusively women) is superimposed into pornographic material without their consent (Adjer et al. 2019).

In a recently reported event, deepfakes have been used to make fake LinkedIn accounts (Satter 2019) to try and connect with employees at the US State Department. As the identity of generated images of faces cannot be found through reverse image search, it makes them perfect for creating false online identities that are difficult to trace as being fake. Unsurprisingly, the production of and means of detecting deepfakes have both become fast growing areas of research, as well as industries in and of themselves (Venkataramakrishnan 2019).

In order to produce deepfakes, the machine learning algorithms that generate these images are optimised in an unsupervised fashion to distinguish between real and generated images. Through this process they become increasingly sophisticated at generating realistic images (see Section 2.1 for more details). Not only are the machines optimised to make representations that are more realistic, they also generate information on whether or not a generated image is fake. To our knowledge, visualising and understanding this aspect of the machine's "disembodied, post-humanized gaze"(Steyerl 2012) has not yet been interrogated. This paper and the series of artworks *Being Foiled*[1] explore the aesthetic outcomes of optimising towards producing images the machine predicts are fake rather than real. By starting from realism and optimising away from it, the process bridges the uncanny valley in reverse, ultimately ending at a point of near total abstraction.

## 2. Background
### 2.1. Machine Learning

Machine Learning is a broad field, closely related to statistics and data mining, but with more of an emphasis on using methods of optimisation to automatically develop programs "that can automatically detect patterns in data ... to predict future data or other outcomes of interest" (Murphy 2012). In the field of Machine Learning, 'learning' is understood as an automated process of optimisation (often referred to as training), where an algorithm processes data and finds a set of parameters that best 'solve' a pre-defined objective function. This paper uses the term 'learning' in this computational

[1]. Artworks from the series can be seen here: https://terencebroad.com/works/being-foiled.

sense, to be considered distinct from the anthropomorphic sense of human understanding, knowledge acquisition and reasoning.

Mackenzie, in his ethnography of Machine Learning, describes the field as a discourse of knowledge-practice not knowledge-consciousness. He describes the 'learning' in machine learning as embodying "a change in programming practice, or indeed the programming of machines ... a contrast between programming as 'a lot of [building] work' and the 'farming' done by machine learners to 'grow' programmes" (2017) .

## 2.2. Generative Adversarial Networks

The key method used to produce deepfakes is the Generative Adversarial Networks (GAN) framework. In this framework there are two main components, the generator that produces random samples and a discriminator that is optimised to classify real data as being real and generated data as being fake. The generator is optimised to try and fool the discriminator. Over time it learns to do so, producing increasingly realistic samples that 'fit' the data distribution without reproducing samples from the dataset.

Since the inception of GANs in 2014 (Goodfellow et al. 2014) a number of breakthroughs have been achieved in improving their fidelity, (Radford, Metz, and Chintala 2016; Karras et al. 2018; Brock, Donahue, and Simonyan 2019) leading to StyleGAN (Karras, Laine, and Aila 2019) which was most likely used to make the fake LinkedIn profiles referred to earlier and acts as the base pre-trained model for this paper.

GANs have also been used widely in the production of art, leading some to declare them leading to the next great movement in art (Schneider and Rea 2018). However, GAN generated art has come under a substantial amount of criticism. The banal production of new artworks, trained on datasets of existing artworks (Christie's 2018) has led Hassine and Ziv to term some examples of GAN generated art as 'zombie art' (2019). Another criticism is the reliance on the ability of deep neural networks to produce endless variations of mesmerising samples, without any meaningful framing of the works being presented by the artists (Zylinska 2019).

## 2.3. Fine-tuning GANs

Once a GAN is trained, the discriminator is usually discarded and the samples are drawn solely from the generator. However, this discriminator network contains potentially powerful representations that can be used in subsequent sampling or fine-tuning procedures of the generator. In our previous work, we showed that by freezing the weights of the discriminator, it can be used in conjunction with features from another network in order to fine-tune the weights of the generator through backpropagation. This enables it to transform its output distribution to a novel distribution with unexpected characteristics (Broad and Grierson 2019a).

## 2.4. The Uncanny

The uncanny is a psychological or aesthetic experience that can be characterised as observing something familiar that is encountered in an unsettling way. In his 1906 essay, Ernst Jentsch defined the uncanny as an experience that stems from uncertainty, giving an example of it as being most pronounced when there is "doubt as to whether an apparently living being is animate and, conversely, doubt as to whether a lifeless object may not in fact be animate" (Jentsch 1997). Freud later refined this definition to argue that the uncanny occurs when something familiar is alienated, when the familiar is viewed in an unexpected or unfamiliar form (1919).

In art, feelings of the uncanny are often evoked to explore boundaries between what is living and what is machine. This often reflects the anxieties and technologies of any given era, such as interactive robotic installations in the late 20th Century (Tronstad 2008). In work from the early 20th Century, such as Jacob Epsteins *Rock Drill* which depicts the human form as transformed and amalgamated by industrial machinery (Grenville 2001). In moving image, Czech animator Jan Svankmajer is well known for creating animated representations of the human form that deliberately confuse the viewer with respect to notions of life and lifelessness (Chryssouli 2019).

## 2.5. The Uncanny Valley

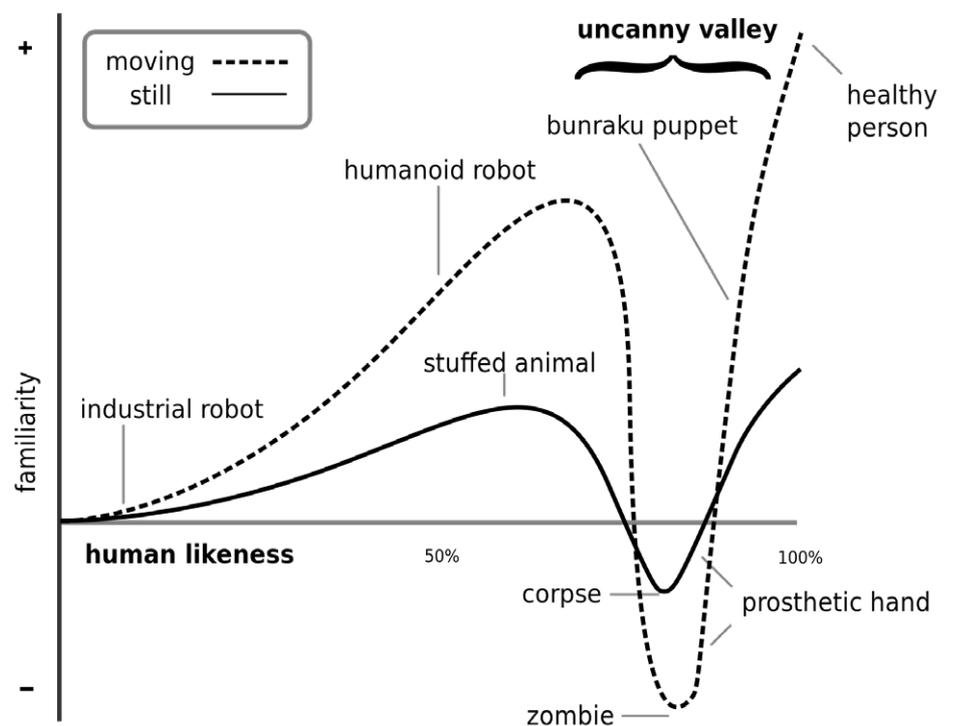

**Fig. 1.** Diagram illustrating the uncanny valley effect. Source: Wikipedia[2] (CC BY-SA 3.0).

**2.** https://simple.wikipedia.org/wiki/Uncanny_valley#/media/File:Mori_Uncanny_Valley.svg

The uncanny valley is a concept first introduced by Masahiro Mori in 1970. It describes how in the field of robotics, increasing human likeness increases feelings of familiarity up to a point (see Figure 1), before suddenly decreasing. As a humanoid robot's representation approaches a great closeness to human form, it provokes an unsettling feeling, and the responses in likeness and familiarity rapidly become more negative than at any prior point. It is only when the robotic form is close to imperceptible with respect to human likeness that the familiarity response becomes positive again (Mori, MacDorman, and Kageki 2012). In addition to being observed in robotics, this has also been observed in CGI, games and other domains where the likeness of humans and animals is imitated.

## 3. Inverting the Objective Function

In similar fashion to our previous work (Broad and Grierson 2019a), we take a pre-trained GAN (in this case StyleGAN trained on the Flickr Faces HQ dataset[3]) and then fine-tune the weights of the generator whilst keeping the weights of the discriminator frozen. The main difference here is that instead of using the discriminator in its standard usage (to determine if a generated sample looks realistic), we invert this objective function, optimising the generator to begin producing images the discriminator sees as being fake.

### 3.1. Maximising Unlikelihood

GANs, in their adversarial game of deception and detection, implicitly learn to maximise the likelihood of generating samples that fit the distribution of the dataset they are given. By inverting this objective function we are, in effect, maximising the unlikelihood of the data.

As we are starting from a pre-trained model, the initial state is a generator that produces highly realistic samples. But as the fine-tuning process occurs, the weights of this model begin to change in accordance with features that are predicted by the network as tell-tale signs that the sample is fake, increasingly exaggerating these features until they are prominent.

## 4. Divergence > Convergence > Collapse

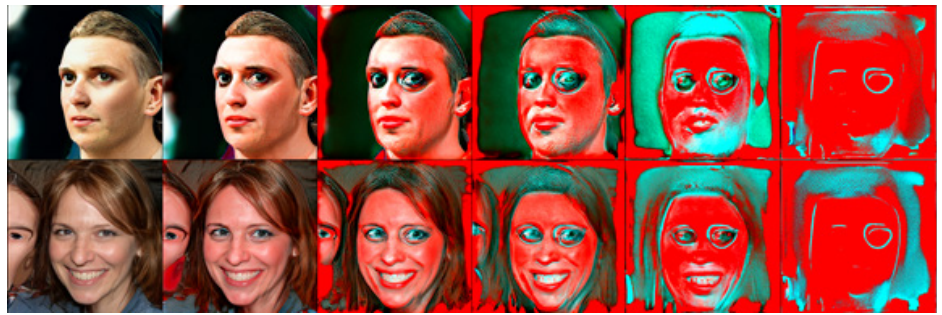

**3.** In this case we use an unofficial PyTorch implementation of StyleGAN and accompanying pre-trained models: https://github.com/rosinality/style-based-gan-pytorch.

**Fig. 2.** Samples from fine-tuning process after 0, 250, 500, 750, 1000 and 1500 iterations.

Over the course of the fine-tuning procedure, the generator goes through a number of stages, increasing in its unlikelihood and in effect, reversing into the uncanny valley. Beginning from a state of producing almost imperceptibly realistic images, to increasingly exaggerating features that show the images are fake, ultimately collapsing into complete abstraction.

The different stages we describe are not discrete, but overlap, interact and feed into one another. We categorise them as representing three prominent phenomena. The first is divergence; the generator slowly starts to diverge from the original distribution of natural images, towards a constantly evolving new distribution, increasing in their uncanniness as the 'fake' or unnatural features of the images become increasingly pronounced.

Secondly (and concurrently), the generator begins to converge towards a new state that maximes the unlikeliness of the data. As this process continues, the gradients (the derivatives of the loss function being back-propagated through the generator) begin to explode. The system is in a vicious cycle where each update to the generator causes it to produce results that the discriminator predicts more strongly as being fake, producing an even more extreme loss function, causing ever more extreme changes to be made to the updates of the parameters of the generator.

This ultimately leads to collapse. The increasingly extreme gradients have washed away any of the subtle or delicate features that were present in the original data. The entire space of potential images has collapsed into (effectively) a single output (see Figure 3), a posterised caricature where human features are barely registrable.

The process we have described is a positive feedback loop, a reinforcement of the intensification of prediction as viewed and understood through the samples generated by the model, giving us a view into the stages of transfiguration it goes through. This process can be thought of as a computational enactment of the production of prediction (Mackenzie 2015), providing a form of visual insight into the production of deepfakes themselves.

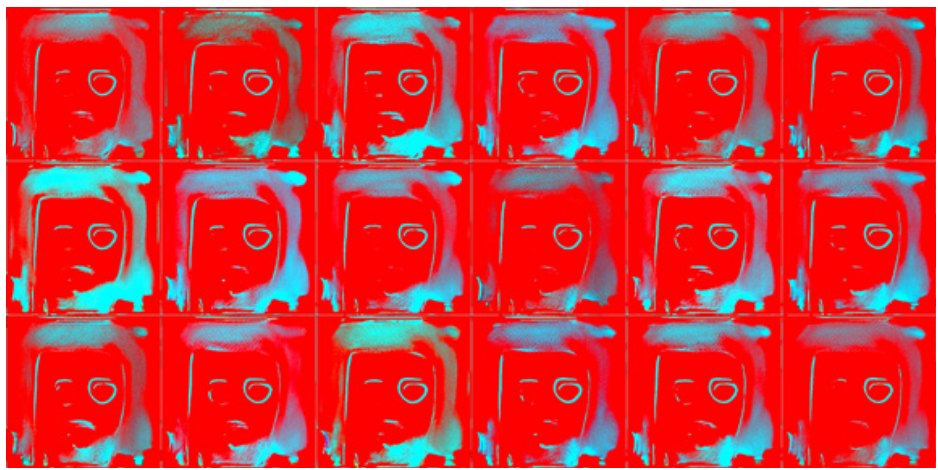

**Fig. 3.** Samples drawn from the model after 2000 iterations (part of the series of works *Being Foiled*).

## 5. Examining Peak Uncanniness

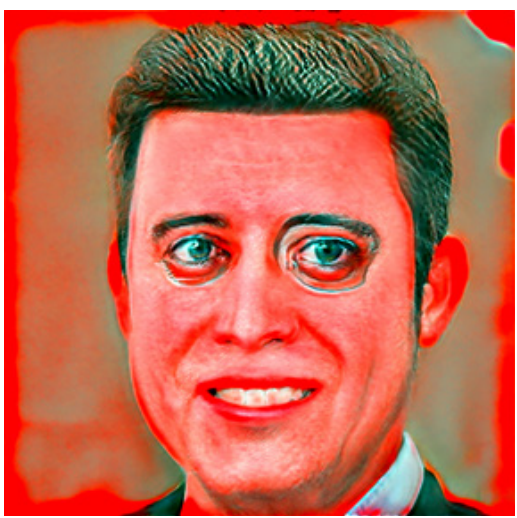

Fig. 4. A sample drawn from the model after 500 iterations (Part of the series of works *Being Foiled*).

If we take a snapshot of the model at an earlier iteration then we can draw samples when (in our opinion) the uncanniness is most pronounced (See Figure 4). This optimisation process has bridged the uncanny valley in reverse, starting from a state where samples are almost imperceptibly life-like, towards almost complete abstraction. By stopping early, we can pick an iteration of the model where the uncanniness is potentially most amplified.

When examining these samples, what is particularly striking is the prominent red hue that has saturated the entire face of the subject, so much so it is bleeding into the surrounding background of the image. This is in stark contrast with the overt blue shades infecting the eyes and peripheral facial regions.

The exaggerated artifacts around the eyes are instructive of the fact that this must be where flaws in the generation most often occur, potentially because there are a lot of details and a wide range of diversity in those details that have to be modelled to produce both realistic faces and an array of distinct identities. The eyes in many of these samples are not aligned, and there is an exaggerated definition around the wrinkles where the eyes are set. This is also the faultline between outputs where faces have or do not have glasses. If the generator produces a sample that is half-way between wearing and not wearing glasses this would be a telltale sign that the image is fake.

There are overt regularities in the texture of the hair. An artifact of the network generating these images from spatial repeated, regular features, and again, something that is a tell-tale sign that the image is generated by a machine. These regularities, asymmetries, and other pronounced artifacts fit with many of the descriptions given by McDonald in his blog post detailing ways of recognizing deepfakes (2018).

**Fig. 5.** Samples drawn from the model after 500 iterations (part of the series of works *Being Foiled*).

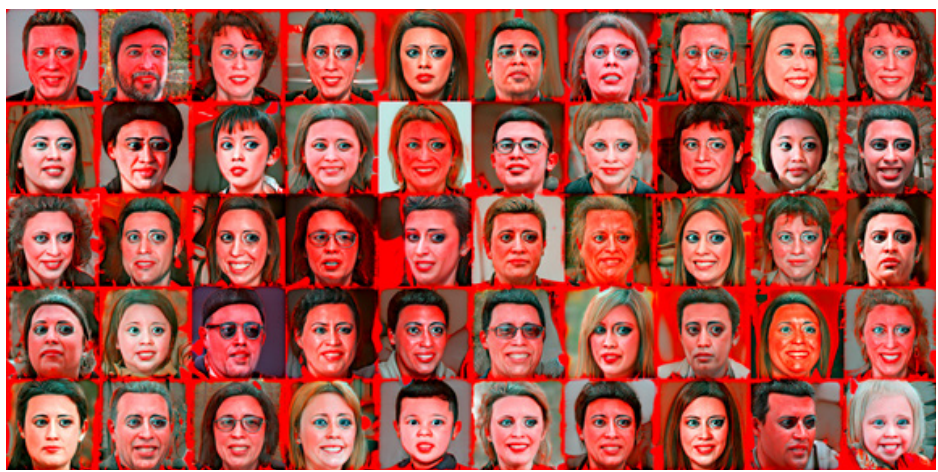

Viewing the samples individually provokes a certain feeling of uncanniness. But when viewed in aggregate across a population of samples (see Figure 5) this feeling is intensified further, provoking (in our view) an almost visceral response, as if viewing a diseased population. Even the emotional register appears off. Many of the samples appear to be half grimacing, having either a completely vacant stare, or one with unnerving intensity.

One criticism of GAN generated art is that the endless generation of samples from a given model, while initial mesmerising and transfixing, can quickly become banal, monotonous repetitions for the sake of overwhelming the viewer with the "sublime of algorithmic productivity" (Zylinska 2019). However we believe that presenting the work in this context helps to provoke and reinforce the feeling of uncanniness in the viewer (just as the intensification of these artifacts across the distribution of generated samples reinforces the prediction of fakeness by the system itself). Therefore, we think it is fitting to present the work as grid(s) of endless variations, all with the same common transformations and transfigurations. The faultlines of fakery emerging from the foreground and the background, bleeding into each other, are an artifact which was purely accidental, but welcomed. An 'accidental aesthetic' that is the result of non-human agency and inter-object relations (Koltick 2015).

## 6. Exploring Different States

In the previous two sections, we discussed the results from one iteration of the model (after training is completed at the resolution 512x512). However, this fine-tuning process can be done at any iteration of the model and seemingly with widely varying results (see Figures 6 & 7).

To understand why there is such variation in the results of this process at different stages of model training, it may be instructive to consider the unusual nature of GAN training. Unlike most machine learning systems, which are most often highly non-linear convex optimisation problems, attempting to find an optimal set of parameters to clearly defined objective functions, GANs operate more like a dynamic system, with no target end state. The optimisation problem is almost circular (Nagarajan and Zico Kolter 2017). The generator and discriminator will endlessly be playing this game of forger/detective. The discriminator endlessly picks up on new miniscule flaws in the generator output, and the generator in turn responds.

With there being no target end state, the flaws most prominent to the discriminator are ever shifting and evolving over the training process. The samples in Figure 6 show that the fine-tuning process pushes the output towards increasingly muddy, washed out images, the facial features, dispersing as if being propagated by waves. In contrast the samples in Figure 7 show a hardening of the facial features. With rectangular geometric regularities in the shape of the nose and mouth becoming increasingly prominent.

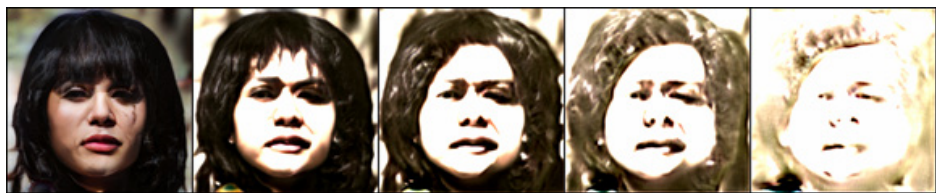

**Fig. 6.** Samples from the fine-tuning process from an earlier iteration (256x256) of the model (after 0, 150, 300, 450, 600 iterations).

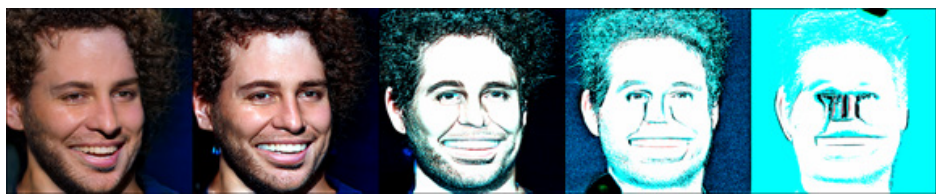

**Fig. 7.** Samples from the fine-tuning process from a later iteration (512x512) of the model (after 0, 150, 300, 450, 600 iterations).

## 7. Conclusion

What makes deepfakes so mesmerising (and terrifying) is our inability to distinguish them from *real* images. What we have done here is to manipulate the weights of a system that would normally produce indistinguishably real images to instead produce definitively unreal images. This we achieve by utilising a key and otherwise unseen component of the predictive capacity of a deepfake system, in the service of producing—the prediction of—unreal images. These are generated by a process that can be understood as a feedback loop that bridges the uncanny valley in reverse, exposing the fragility and arbitrary nature of the configuration of parameters that produces such realistic images, whilst showing that it can quickly be twisted and exposed to produce distinctly unreal, previously unseen, images.

Comparing the results of this process from different iterations of the model, it is apparent that what the machine predicts as being fake is constantly in flux. The relationship between generator and discriminator (forger/detective) is constantly evolving. It is this constantly evolving dynamical relationship which makes GANs so effective at producing realistic deepfakes in the first place.

It is important to note that these results are highly contingent on a number of factors: the idiosyncrasies of the state of the model snapshot that was selected from the training process, the choice of methods for optimisation and regularisation and accompanying hyperparameters used in training (which can have a profound effect on the visual output), the tacit knowledge that has been ascertained through previous experiments of creating novel network ensembles (Broad and Grierson 2019b), and fine-tuning existing models (Broad and Grierson 2019a).

This work fits into a broader line of enquiry of technical and artistic research looking at new ways of configuring ensembles of neural network models, examining the interaction of the models, the interaction of the artist with the models, and the aesthetic outcomes of the generative processes. By manipulating and reconfiguring these generative models, there is a huge amount of latent potential for such systems to produce not only novel outcomes, but learning sets of parameters that can produce novel, *divergent distributions* of outcomes that do not imitate any distribution of images created or curated by people. With the series of artworks *Being Foiled*, we have presented the results of these experiments in a way that can be understood visually, giving a representation of an otherwise unseen and uncanny aspect of the machine's gaze.

**Acknowledgements:** This work has been supported by the UK's EPSRC Centre for Doctoral Training in Intelligent Games and Game Intelligence (IGGI; grant EP/L015846/1).